\documentclass[10pt,twocolumn,letterpaper]{article}

\usepackage{iccv}
\usepackage{times}
\usepackage{epsfig}
\usepackage{graphicx}
\usepackage{amsmath}
\usepackage{amssymb}

\usepackage{url}
\usepackage{booktabs}
\usepackage{multirow}
\usepackage{colortbl}
\usepackage{float}
\usepackage{xcolor}
\usepackage{makecell}

\usepackage{pifont}
\usepackage{soul} % \st
\usepackage{dsfont}
\usepackage[pagebackref=true,breaklinks=true,letterpaper=true,colorlinks,bookmarks=false]{hyperref}

\usepackage[capitalize]{cleveref}  % 并排图
\iccvfinalcopy % *** Uncomment this line for the final submission

\def\iccvPaperID{6158} % *** Enter the ICCV Paper ID here

\ificcvfinal\fi

\begin{document}

%%%%%%%%% TITLE
\title{DETRDistill: A Universal Knowledge Distillation Framework for DETR-families}

\author{Jiahao Chang\textsuperscript{1*} \quad Shuo Wang\textsuperscript{1*} 
\quad Hai-Ming Xu\textsuperscript{2*} \quad Zehui Chen\textsuperscript{1} \quad Chenhongyi Yang\textsuperscript{3} \quad Feng Zhao\textsuperscript{1$\dagger$} \\
\textsuperscript{1}University of Science and Technology of China\\
\textsuperscript{2}University of Adelaide\\
\textsuperscript{3}University of Edinburgh\\
}

\maketitle
\ificcvfinal\thispagestyle{empty}\fi

%%%%%%%%% ABSTRACT
\begin{abstract}

Transformer-based detectors~(DETRs) are becoming popular for their simple framework, but the large model size and heavy time consumption hinder their deployment in the real world.
While knowledge distillation(KD) can be an appealing technique to compress giant detectors into small ones for comparable detection performance and low inference cost.
Since DETRs formulate object detection as a set prediction problem, existing KD methods designed for classic convolution-based detectors may not be directly applicable.
In this paper, we propose \textbf{DETRDistill}, a novel knowledge distillation method dedicated to DETR-families.
Specifically, we first design a Hungarian-matching logits distillation to encourage the student model to have the exact predictions as that of teacher DETRs.
Next, we propose a target-aware feature distillation to help the student model learn from the object-centric features of the teacher model.
Finally, in order to improve the convergence rate of the student DETR, we introduce a query-prior assignment distillation to speed up the student model learning from well-trained queries and stable assignment of the teacher model.
Extensive experimental results on the COCO dataset validate the effectiveness of our approach. Notably, DETRDistill consistently improves various DETRs by more than 2.0 mAP, even surpassing their teacher models.
% The student model with our DETRDistill even outperforms its teacher model between three baselines.

% For example, DETRDistill consistently achieves over 2.0 AP performance gains on the COCO val set when Conditional DETR, Deformable DETR and AdaMixer served as teachers, respectively.

\end{abstract}

%%%%%%%%% BODY TEXT
\section{Introduction}

Object detection aims to locate and classify visual objects from an input image. In the early works, the task was typically achieved by incorporating convolution neural networks (CNNs) to process the regional features of the input image~\cite{ren2015faster,lin2017focal}, in which a bunch of inductive biases were included, such as anchors~\cite{ren2015faster}, label assignment~\cite{zhang2020bridging} and duplicate removal~\cite{bodla2017soft}. Recently, transformer-based object detectors like DETR~\cite{carion2020end} have been proposed where detection is treated as a set prediction task 
which significantly simplifies the detection pipeline and helps the users free from the tedious tuning of the hand-craft components, e.g., anchor sizes and ratios~\cite{ren2015faster}.

\begin{figure}[t]
  \centering
%   \fbox{\rule{0pt}{2in} \rule{0.9\linewidth}{0pt}}
   \includegraphics[width=1\linewidth]{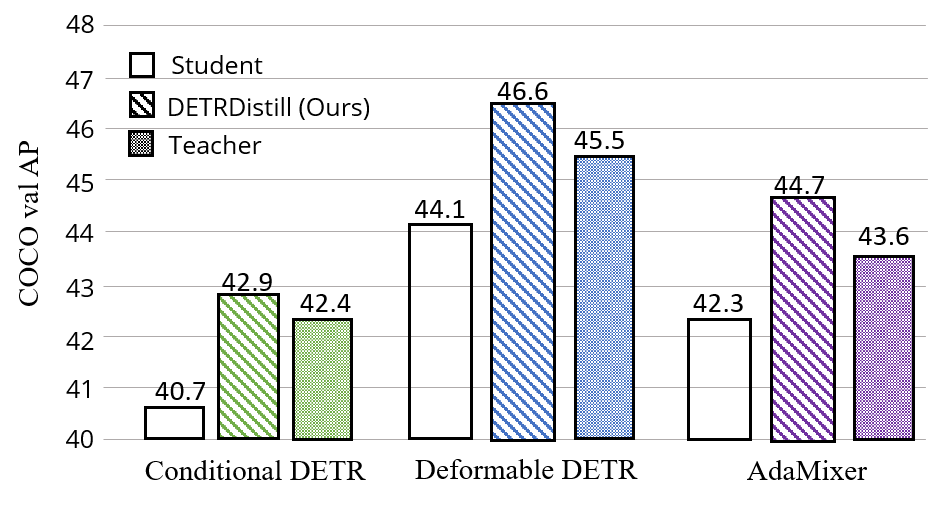}

   \caption{The performance of our DETRDistill on three transformer-based detectors: Conditional DETR~\cite{meng2021conditional}, Deformable DETR~\cite{zhu2020deformable}, and AdaMixer~\cite{gao2022adamixer}. We adopt ResNet-101 and ResNet-50 for the teacher and student models, respectively. Our DETRDistill yields significant improvements compared to its student baseline, and even outperforms its teacher model.}
   \label{fig:performance}
\end{figure}

Although the transformer-based detectors have achieved state-of-the-art performance~\cite{liu2022dabdetr,li2022dn,gao2022adamixer}, they suffer from an expensive computation problem, making them difficult to be deployed in real-time applications. In order to acquire a fast and accurate detector, \textit{knowledge distillation}~\cite{hinton2015distilling} (KD) is an appealing technique. Normally, KD methods transfer knowledge from a heavy-weighted but powerful teacher model to a small and efficient student network by mimicking the predictions~\cite{hinton2015distilling} or feature distributions~\cite{romero2014fitnets} of teachers.

\begin{figure}[t]
  \centering
%   \fbox{\rule{0pt}{2in} \rule{0.9\linewidth}{0pt}}
   \includegraphics[width=1\linewidth]{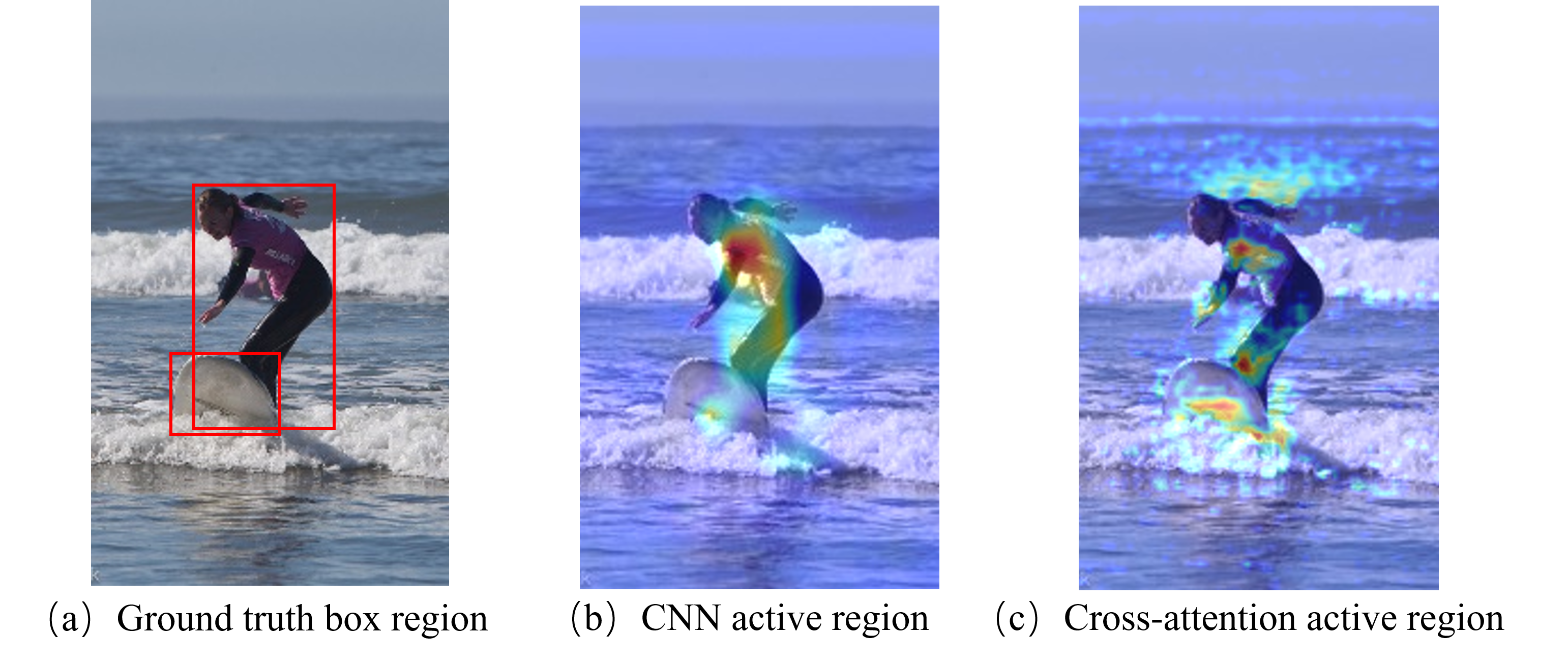}

   \caption{Visualization of the ground truth boxes (a) and active region from ATSS~\cite{zhang2020bridging} (b) and AdaMixer~\cite{gao2022adamixer} (c).}
   \label{fig:gt_cnn_detr}
\end{figure}

In the research area of object detection, there have various kinds of KD methods been published~\cite{yang2022focal,yang2022predictionguided,zagoruyko2016paying,romero2014fitnets,chen2017learning,li2017mimicking,guo2021distilling,zheng2022localization,yang2022masked,touvron2021training,song2021vidt}. However, most of these methods are designed for convolution-based detectors and may not directly apply to transformer-based DETRs due to the detection framework differences. There are at least two challenges: \ding{182} \textit{logits-level distillation methods}~\cite{hinton2015distilling,zheng2022localization} \textit{are unusable for DETRs.} For either anchor-based~\cite{zhang2020bridging} or anchor-free~\cite{tian2019fcos} convolution-based detectors, box predictions are closely related to the feature map grid and thus naturally ensure a strict spatial correspondence of box predictions for knowledge distillation between teacher and student. However, for DETRs, box predictions generated from the decoders are unordered and there is no natural one-to-one correspondence of predicted boxes between teacher and student for a logits-level distillation. \ding{183} \textit{Feature-level distillation approaches may not suitable for DETRs.} Due to the feature generation mechanism being different between convolution and transformer \cite{romero2014fitnets}, the region of feature activation for the interested object varies a lot. As Fig.~\ref{fig:gt_cnn_detr} shows, the active region of a convolution-based detector is almost restricted inside the ground truth box while the DETR detector further activates regions in the background area. Therefore, directly using previous feature-level KD methods for DETRs may not necessarily bring performance gains and sometimes even impair the student detectors as presented in Tab.~\ref{tab:fea_distill}.

\begin{table}[t]
\begin{center}

% \resizebox{\linewidth}{!}{
\begin{tabular}{c|ccc}
  \toprule
  \multirow{2}*{Method} & \#Epoch4 & \#Epoch8 & \#Epoch12 \\
   & AP & AP & AP \\
  \midrule
  \midrule
  Baseline w/o KD &  35.0 & 38.7& 42.3 \\ 
  FGD~\cite{yang2022focal} & 34.4(-0.6) & 39.1(+0.4)& 40.7(-1.6)  \\
  FKD~\cite{zhang2020improve} & 35.9(+0.9) & 39.5(+0.8)& 42.2(-0.1)  \\ 
  MGD~\cite{yang2022masked} & 36.3(+1.3)& 39.8(+1.1)& 42.3(+0.0)  \\ 
  FGFI~\cite{wang2019distilling} & 35.6(+0.6) & 39.3(+0.6) & 42.6(+0.3) \\ 
  FitNet~\cite{romero2014fitnets} & 36.4(+1.4)& 39.6(+0.9) & 42.9(+0.6)  \\ 
  % \midrule
  % Ours & \textbf{38.3}(+3.3)& \textbf{41.0}(+2.3) & \textbf{43.5}(+1.2)  \\ 
  \bottomrule
\end{tabular}
% }
\end{center}
\caption{Comparison on several CNN-based region weighted feature distillation approaches on AdaMixer~\cite{gao2022adamixer}.}
\label{tab:fea_distill}
\end{table}

To address the above challenges, we propose \textit{\textbf{DETRDistill}}, a knowledge distillation framework specially designed for detectors of DETR families. 
Precisely, DETRDistill mainly consists of three components: 

(1) Hungarian-matching logits distillation: To solve the challenge \ding{182}, we use the Hungarian algorithm to find an optimal bipartite matching between predictions of student and teacher and then the KD can be performed in the logits-level.
However, since the number of boxes predicted as positive in the teacher model is quite limited, doing KD only on positive predictions does not bring a significant performance gain. Instead, we propose to introduce a distillation loss on the massive negative predictions between the teacher and student model to fully make use of the knowledge lying in the teacher detector. Moreover, considering DETR methods typically contain multiple decoder layers for a cascade prediction refinement, we also create a KD loss at each stage to have a progressive distillation.

(2) Target-aware feature distillation:
According to the analysis in challenge \ding{183}, we propose utilizing object queries and teacher model features to produce soft activation masks. Since the well-trained teacher queries are closely related to various object targets, such generated soft masks will be object-centric and thus make soft-mask-based feature-level distillation to be target-aware.

(3) Query-prior assignment distillation: 
Since the queries and decoder parameters are randomly initialized in the student model, unstable bipartite assignment in the student model leads to a slow convergence rate as presented in~\cite{li2022dn}. While we empirically find that well-trained queries in the teacher model can always produce a consistent bipartite assignment as shown in Fig.~\ref{fig:stability}, we thus propose to let the student model take teacher's queries as an additional group of prior queries and encourage it to produce predictions based on the stable bipartite assignment of the teacher network. Such a proposed distillation successfully helps the student model to converge fast and achieve better performance.

In summary, our contributions are in three folds:

$\bullet$ We analyze in detail the difficulties encountered by DETRs in the distillation task compared with traditional convolution-based detectors.

$\bullet$ We propose multiple knowledge distillation methods for DETRs from the perspective of logits-level, feature-level, and convergence rate, respectively.

$\bullet$ We conduct extensive experiments on the COCO dataset under different settings, and the results prove the effectiveness and generalization of our proposed methods.
%-------------------------------------------------------------------------

\begin{figure*}
  \centering
   \includegraphics[width=0.8\linewidth]{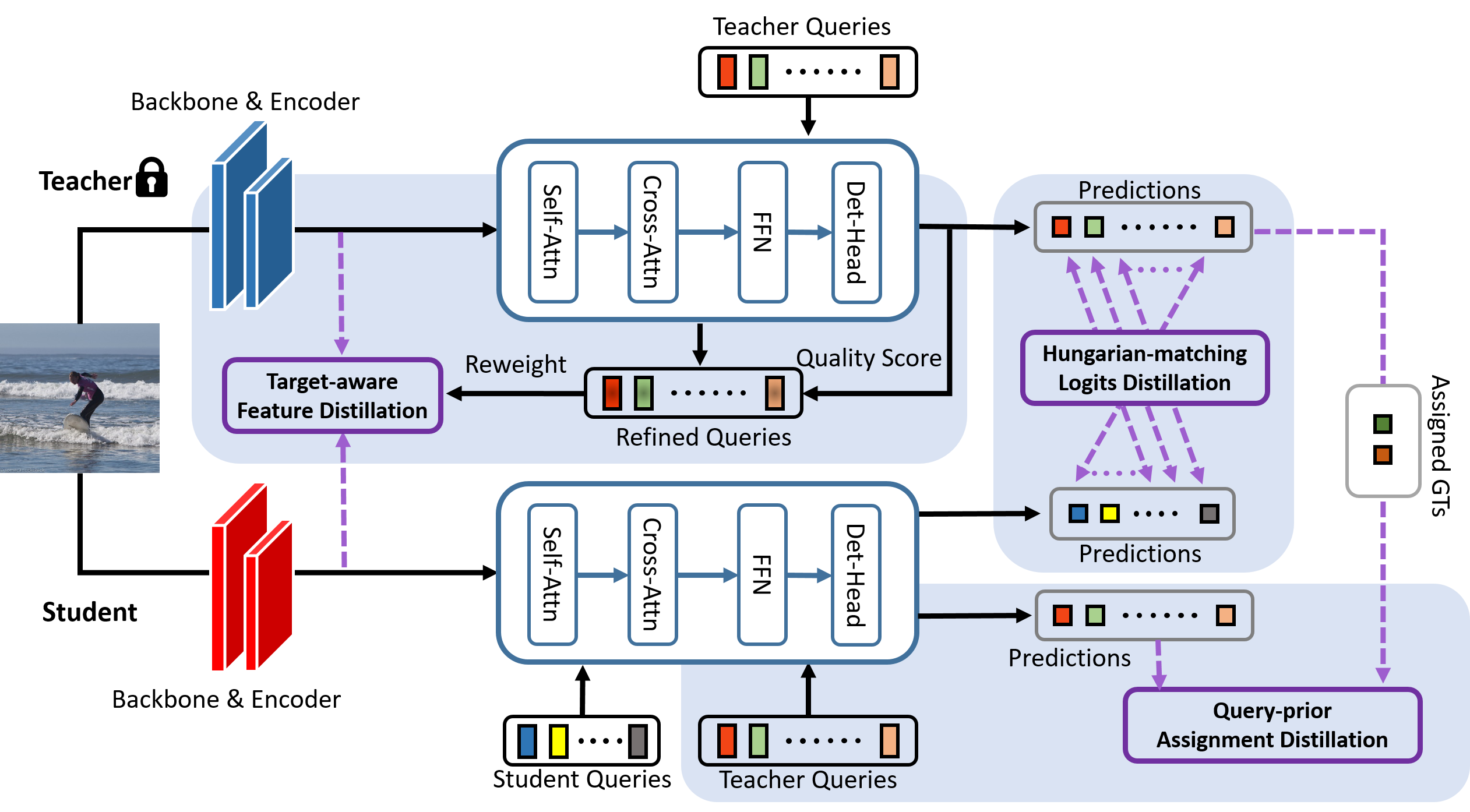}
   \caption{The overall architecture of our approach, consisting of a transformer-based teacher detector with a large backbone, a congener detector with a lightweight backbone, and the proposed distillation modules: (i) Hungarian-matching Logits Distillation (ii)  Target-aware Feature Distillation, and (iii) Query-prior Assignment Distillation. We omit the original training supervision for clear formulation.}
   \label{fig:main_fig}
\end{figure*}

\section{Related Work}
\label{sec:formatting}

\subsection{Transformer-based Object Detectors}
With the excellent performance of Transformer~\cite{vaswani2017attention} in natural language processing, researchers have also started to explore the application of Transformer structure to visual tasks~\cite{liu2021swin, chen2021visformer, mao2022towards}. However, the DETR training process is extremely inefficient, so many follow-up works have attempted to accelerate convergence. One line of work tries to redesign the attention mechanism. For example, Dai \etal~\cite{zhu2020deformable} propose Deformable DETR, which constructs a sparse attention mechanism by only interacting with the variable sampling point features around the reference points. SMCA~\cite{gao2021fast} introduces Gaussian prior to limit cross-attention. AdaMixer~\cite{gao2022adamixer} designs a new adaptive 3D feature sampling strategy without any encoder and then mixes sampled features in the channel and spatial dimension with adaptive weights. 

Another line of work rethink the meaning of the query. Meng \ \etal~\cite {meng2021conditional} visualize that it is ineffective for DETR to rely on content embedding in the cross-attention to locate object extremity and therefore propose decoupling queries into the content part and position part. Anchor-DETR~\cite{wang2022anchor} directly treats the query's 2D reference points as its position embedding to guide attention. DAB-DETR~\cite{liu2022dab} introduces width and height information besides location to the attention mechanism to model different scale objects. DN-DETR~\cite{li2022dn} introduces the query denoising task to accelerate the training. Group-DETR~\cite{chen2022group} and H-DETR~\cite{jia2022detrs} improve performance by increasing positive samples as auxiliary groups in the decoder training. Unlike the previous work, we expect to improve the performance of small models through distillation.

\subsection{Knowledge Distillation in Object Detection}

Knowledge distillation is a commonly used method for model compression. \cite{hinton2015distilling} first proposes this concept and applies it in image classification. They argue that soft labels output by the teacher contains ``dark knowledge'' of inter-category similarity compared to the one-hot encoding, which contributes to the model's generalization. Attention transfer~\cite{zagoruyko2016paying} focuses the distillation on the feature map and transfers knowledge by narrowing the attention distribution of the teacher and student instead of distilling output logits. 

FitNet~\cite{romero2014fitnets} proposes mimicking the teacher model's intermediate-level hints by hidden layers. \cite{chen2017learning} first applies knowledge distillation to solve the multi-class object detection. \cite{li2017mimicking} thinks that the background regions would introduce noise and proposes distilling the regions RPN sampled. DeFeat~\cite{guo2021distilling} distills the foreground and background separately. FGD~\cite{yang2022focal} imitates the teacher regarding focal regions and global relations of features, respectively. LD~\cite{zheng2022localization} extends soft-label distillation to positional regression, causing the student to fit the teacher's border prediction distribution. MGD~\cite{yang2022masked} uses masked image modeling~(MIM) to transform the imitation task into a generation task. In addition to the above CNN-based distillation, some works involve vision transformers. DeiT~\cite{touvron2021training} transfers inductive biases into ViT~\cite{dosovitskiy2020image} from a CNN-teacher through a distillation token and achieves competitive performance in the classification task. ViDT~\cite{song2021vidt} performs KD on patch tokens and proposes a variation of the transformer detector. However, such a distillation can not apply to the DETR-families directly. Our work analyzes the unique phenomena of different components of DETR and proposes a universal distillation strategy.
%-------------------------------------------------------------------------

\section{A Review of DETR}

DETR~\cite{carion2020end} is an end-to-end object detector that includes a backbone, Transformer encoders, learnable query embeddings and decoders. Given an image $I$, the CNN backbone extracts its spatial features and then Transformer encoders~(some variants do not require encoders~\cite{gao2022adamixer}) will enhance the feature representation. With the updated features $F \in R^{HW \times d}$, query embeddings $Q \in R^{N \times d}$ are fed into several Transformer (typically six) decoders, where $d$ is the feature dimension, $N$ is the fixed number of queries. The operations in each decoder stage are similar: Firstly, exploiting self-attention to establish the relationship between queries to capture the mutual information. Secondly, interacting queries with image features via flexible cross-attention to aggregate queries with valuable semantic information. Thirdly, a feed-forward network (FFN) decodes each query into $\hat{y}_i=(\hat{c}_i, \hat{b}_i)$ including predicted categories and bounding boxes. 

In the training stage, the principle of label assignment is to minimize the matching cost between model prediction and ground truth (GT) to get a bipartite matching with Hungarian algorithm~\cite{kuhn1955hungarian}. The optimal matching is solved as follows:
\begin{equation}\label{1}
\hat{\sigma} = arg\,min_{\sigma}\sum_{i=1}^N \mathcal{L}_{\mathrm{match}}(y_{i}, \hat{y}_{\sigma_i}), \nonumber
\end{equation}
where $\sigma_i$ is a permutation of $N$ elements and $\hat{\sigma}$ is the optimal assignment. $y_{i}=(c_{i}, b_{i})$ is $i$-th GT where $c_{i}$ is the target class (which may be $\varnothing$) and $b_{i}$ is the GT box. $\mathcal{L}_{\mathrm{match}}$ is a pair-wise matching cost:
\begin{equation}\label{2}
\mathcal{L}_{\mathrm{match}}(y_{i}, \hat{y}_{\sigma_i}) = \mathcal{L}_{\mathrm{cls}}(c_i, \hat{c}_{\sigma_i}) + \mathds{1}_{\left\{c_{i} \neq \varnothing\right\}} \mathcal{L}_{\mathrm{bbox}}(b_i, \hat{b}_{\sigma_i}),\nonumber
\end{equation}
where $\mathcal{L}_{\mathrm{cls}}$ and $\mathcal{L}_{\mathrm{bbox}}$ denote the classification and bounding-box losses respectively. Therefore, each GT in the DETR will correspond to only one positive sample query, and all the rest queries are seen as negative samples. The final detection loss function is defined as :
\begin{equation}\label{eq:det}
\mathcal{L}_{\mathrm{det}}(y, \hat{y}_{\hat{\sigma}}) = \sum_{i=1}^N \mathcal{L}_{\mathrm{match}}(y_{i}, \hat{y}_{\hat{\sigma}_i}).
\end{equation}
where the location regression of negative samples is not subject to any supervision.
%-------------------------------------------------------------------------

\section{Our Approach: DETRDistill}

In this section, we introduce the details of our proposed DETRDistill, which consists of three components: (1) Hungarian-matching Logits Distillation; (2) Target-aware Feature Distillation; (3) Query-prior Assignment Distillation. Fig.~\ref{fig:main_fig} shows the overall architecture of DETRDistill.

\subsection{Hungarian-matching Logits Distillation\label{sec:4.1}}

One of the most common strategies for knowledge distillation is to directly align the predictions at the logits-level between the two models. However, query-based predictions in a set form~\cite{li2022dn} make it difficult for DETRs to orderly correspond the teacher's results to that of the student. To achieve this goal, we reuse the Hungarian algorithm to match the predictions from the teacher to that of the student one-to-one.

Formally, let $\hat{y}^T$ and $\hat{y}^S$ denote the predictions from the teacher and student model, conforming to $\hat{y}^T=\bigl\{\{\hat{y}^{T\_pos}_{i}\}_{i=1}^{M^{pos}}, \{\hat{y}^{T\_neg}_{j}\}_{j=1}^{M^{neg}}\bigr\}$ and $\hat{y}^S=\{\hat{y}^{S}_{i}\}_{i=1}^N$, where $M^{pos}$ and $M^{neg}$ denote the number of positive and negative predictions of teacher model. $M=M^{pos}+M^{neg}$ and $N$ are the total numbers of decoder queries for the teacher and student, respectively. $M$ is usually greater than or equal to $N$. Since teacher's positive predictions are target closely related, one straightforward idea is to treat them as knowledgeable pseudo GTs and utilize the Hungarian algorithm to find a matching $\hat{\sigma}^{pos}$ between these positive predictions $\hat{y}^{T\_pos}$ and the ones of student $\hat{y}^S$. Then the logits-level KD can be achieved 
\begin{equation}\label{eq:logitKD_pos}
\mathcal{L}_{\mathrm{logitsKD}}^{pos}(\hat{y}^{T\_pos}, \hat{y}_{\hat{\sigma}^{pos}}^S) = \sum_{i=1}^N \mathcal{L}_{\mathrm{match}}(\hat{y}^{T\_pos}_{i}, \hat{y}_{\hat{\sigma}_i^{pos}}^S).
\end{equation}

However, we empirically find that such a naive KD only brings minor performance gains as presented in Tab.~\ref{tab:pos_neg}. We postulate that the number of positive predictions is quite limited (only 7 per image on average while the total number of queries usually exceeds 100) and the distilled information highly coincides with GTs. On the other hand, a large number of negative predictions of the teacher model are ignored and we argue that these predictions are valuable.

\noindent\textbf{Negative Location Distillation: }Since the teacher model is usually well-optimized, the generated positive predictions and the negative ones may have an obvious difference so that the Hungarian algorithm can produce a plausible assignment, \textit{i.e}., these negatively predicted boxes will stay off the object targets. While the randomly initialized student network may not have such an effect and the student's negative predictions may entangle with positive ones. Therefore, we propose to create a distillation to leverage the knowledge lying in the teacher's negative predictions.\footnote{Since the classification in the teacher's negative predictions may not necessarily always be predicted as background, KD is only done on boxes.}

\begin{equation}\label{eq:logitKD_neg}
\mathcal{L}_{\mathrm{logitsKD}}^{neg}(\hat{y}^{T\_neg}, \hat{y}_{\hat{\sigma}^{neg}}^S) = \sum_{i=1}^N \mathcal{L}_{\mathrm{match}}(\hat{y}^{T\_neg}_{i}, \hat{y}_{\hat{\sigma}_i^{neg}}^S),
\end{equation}
where $\hat{\sigma}^{neg}$ denotes the assignment between $\hat{y}^{T\_neg}$ and $\hat{y}^S$.

\noindent\textbf{Progressive Distillation: }Considering DETR decoders typically contain multiple stages and stage-wise supervision is included by default~\cite{carion2020end}, we further propose to introduce the KD losses in Eq.~\ref{eq:logitKD_pos} and ~\ref{eq:logitKD_neg} into each decoder stage to have a progressive distillation
\begin{equation}\label{eq:logitKD}
    \mathcal{L}_{\mathrm{logitsKD}} 
 = \sum_{k=1}^K \mathcal{L}_{\mathrm{logitsKD}}^{pos}[k] + \mathcal{L}_{\mathrm{logitsKD}}^{neg}[k]
\end{equation}
where $K$ is the number of decoder stages. $\mathcal{L}_{\mathrm{logitsKD}}^{pos}[k]$ and $\mathcal{L}_{\mathrm{logitsKD}}^{neg}[k]$ denote the positive KD loss and negative KD loss on $k$-th decoder stage respectively. 

Please note that we transfer knowledge from the teacher model's stage-by-stage outputs to the student model's corresponding stage, but not simply using the teacher's last stage output to supervise all stages of the student model. It is because we think the teacher model will contain different knowledge at different stages as observed in the recent work~\cite{chen2022enhanced} and the former distillation strategy can make full use of the knowledge in the teacher model and the empirical results in Tab.~\ref{tab:progressive} verify our argument.

% this progressive distillation
\subsection{Target-aware Feature Distillation}
\label{sec:feat_distill}
Detection performance is highly determined by feature representations generated from the feature pyramid network (FPN),
% the backbone and transformer encoders, 
this can be attributed to their rich semantic information related to object targets. Therefore, we argue that distilling the teacher model's knowledge at the feature-level is necessary. The typical manner to mimic the spatial features of a teacher model can be calculated as 
\begin{equation}\label{5}
\mathcal{L}_{\mathrm{featKD}}=\frac{1}{d H W} \left\|\psi \odot \left(\boldsymbol{F}^{T}- \phi(\boldsymbol{F}^{S})\right)\right\|_{2}^{2},
\end{equation}
where $\boldsymbol{F}^{T}\in R^{H\times W\times d}$ and $\boldsymbol{F}^{S}\in R^{H\times W\times d^S}$ denote feature representations produced by the teacher and student model, respectively. $H$ and $W$ denote the height and width of the feature and $d$ is the channel number of teacher's feature.
$\phi$ is a learnable dimension adaptation layer to transform the student's feature into $d$ dimension. $\odot$ is the Hadamard product of two matrices. $\psi \in \mathbb{R}^{H \times W}$ denotes a soft mask for the selection of knowledgeable regions in various KD methods, \textit{e.g}., Romero \etal~\cite{romero2014fitnets} treat the mask as a matrix filled with 1. Wang \etal~\cite{wang2019distilling} generate the mask based on the IoU scores between anchor boxes and GT ones. Sun \etal~\cite{sun2020distilling} utilize the Gaussian mask to cover the GT boxes. Different from the above approaches, we propose to construct the soft mask for DETRs by calculating the similarity matrix between query embeddings and feature representations. Formally, given a whole set of queries $\boldsymbol{Q}^{T} \in R^{M \times d}$ of the teacher model, the selection mask can be obtained

\begin{equation}\label{6}
\psi =\frac{1}{M}\sum_{i=1}^{M} \psi_i, ~~ \mathrm{where} ~~ \psi_i = \boldsymbol{F}^T \cdot \boldsymbol{Q}_{i}^{T}, 
\end{equation}
where $\boldsymbol{Q}_{i}^T\in R^{1\times d}$ is the $i$-th teacher's query and $M$ denotes the number of decoder queries of the teacher's model. 

\begin{figure}[t]
  \centering
   \includegraphics[width=0.9\linewidth]{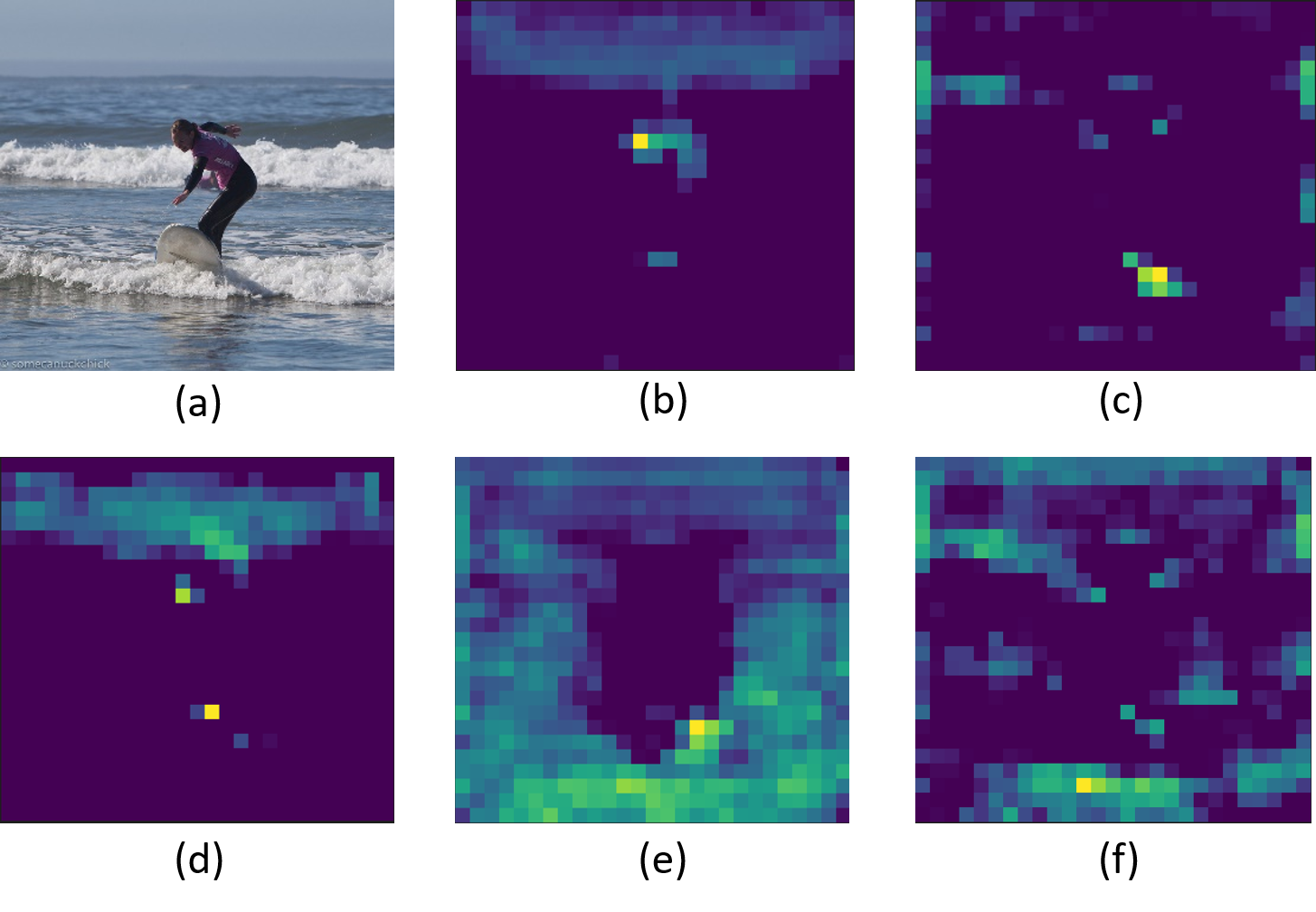}

   \caption{Visualization of the attention mask of queries. (a) is the original image, (b) and (c) are the attention masks generated by the queries responsible for the human and surfboard prediction, respectively. The query corresponding to (d) relates to both the human and the surfboard. (e) and (f) are the masks generated by the samples with low-quality scores.}
   \label{fig:query_mask}
\end{figure}

However, we empirically find such a vanilla distillation approach works poorly as shown in Tab.~\ref{tab:feaKD_ablation}. We postulate the reason is that not all object queries of the teacher should be equally treated as valuable cues. Based on the predictions generated from the teacher queries, Fig.~\ref{fig:query_mask} presents some visualizations of query-based masks $\{\psi_i\}$ and we can find that masks with low-prediction scores attend out of the object regions.

According to the observation, we propose to utilize teacher queries selectively for generating the mask $\psi$. Specifically, we use the quality score proposed in~\cite{chen2021disentangle} as the measurement
\begin{equation}\label{eq:quality}
q_{i}=\left(c_{i}\right)^{\gamma} \cdot \operatorname{IoU}\left(b_{i}^{GT}, b_{i}^{\text {pred}}\right)^{1-\gamma},
\end{equation}
where $c_{i}$ and $b_{i}^{\text{pred}}$ denote the classification score and the predicted box from the $i$-th teacher's query, respectively. $b_i^{GT}$ is the corresponding bipartite matched GT box. $\gamma=0.5$ is a hyperparameter for balancing the weight of classification scores and box IoUs.
Then the target-aware quality score serves as an indicator to guide which query should contribute more to the knowledge distillation and the KD loss in Eq.~\ref{5} can be extended as
\begin{equation}\label{eq:featKD}
\footnotesize
\mathcal{L}_{\mathrm{featKD}}=\sum_{i=1}^{M}\frac{q_{i}}{M d H W} \left\|\psi_{i} \odot \left(\boldsymbol{F}^{T}- \phi\left(\boldsymbol{F}^{S}\right)\right)\right\|_{2}^{2}.
\end{equation}

\begin{table}[t]
\begin{center}

\resizebox{\linewidth}{!}{
\begin{tabular}{c|cccc}
  \toprule
  Setting & AP  & AP$_S$ & AP$_M$ & AP$_L$ \\
  \midrule
  \midrule
  Baseline w/o KD &  42.3 & 25.3& 44.8 & 58.2 \\ 
  % KD w/o quality score & 41.9(-0.4) & 24.4 & 44.9 & 57.4 \\
  KD w/ Eq.~\ref{6} & 41.9(-0.4) & 24.4 & 44.9 & 57.4 \\
  % Target-aware KD & \textbf{43.5}(+1.2)& \textbf{25.4} & \textbf{46.7} & \textbf{60.0}  \\ 
  KD w/ Eq.~\ref{eq:featKD} & \textbf{43.5}(+1.2)& \textbf{25.4} & \textbf{46.7} & \textbf{60.0}  \\ 
  \bottomrule
\end{tabular}
}
\end{center}
\caption{Ablation on feature-level distillation.}
\vspace{6mm}
\label{tab:feaKD_ablation}
\end{table}

\subsection{Query-prior Assignment Distillation}
\label{sec:query_prior_assign_distill}
Since the queries and decoder parameters in DETR are normally randomly initialized for model optimization, a query may be assigned to different objects in different training epochs and thus lead to an unstable bipartite graph matching and a slow convergence speed~\cite{li2022dn}. In the setting of KD, the training of student DETR also suffers from the same problem. However, we empirically observe that the well-optimized queries in the teacher model can consistently achieve a stable bipartite assignment between different decoder stages as presented in Fig.~\ref{fig:stability} and it is intuitive to utilize the knowledge of the teacher model to improve the stability of the training of the student model. Based on this motivation, we propose Query-prior assignment distillation.

Specifically, given teacher query set $\boldsymbol{Q}^{T}$, and we can obtain the corresponding assignment permutation $\hat{\sigma}^{T}$ from the teacher for any given input-GT pairs. We propose to input the teacher query embeddings $\boldsymbol{Q}^{T}$ into the student model as an additional group of prior queries and directly use the teacher's assignment $\hat{\sigma}^{T}$ to generate detection results for loss calculation
\begin{equation}\label{eq:assignKD}
\mathcal{L}_{\mathrm{assignKD}}(y, \hat{y}^S_{\hat{\sigma}^T_i}) = \sum_{i=1}^M \mathcal{L}_{\mathrm{match}}(y_{i}, \hat{y}^S_{\hat{\sigma}^T_i}).
\end{equation}

This proposed KD loss will help the student model treat the teacher queries as a prior and encourage the student detector to achieve a stable assignment as much as possible. As shown in Fig.~\ref{fig:stability}, the matching stability of the student model has been greatly improved with the proposed distillation loss. Please note that such an additional teacher query group is only used during training and the student model will use its default query set for final evaluation.

\subsection{Overall loss}
To sum up, the total loss for training the student DETR is a weighted combination of Eq.~\ref{eq:det}, Eq.~\ref{eq:logitKD}, Eq.~\ref{eq:featKD}, and Eq.~\ref{eq:assignKD}:
\begin{equation}\label{8}
\mathcal{L}= \mathcal{L}_{\mathrm{det}} + \lambda_{1}\mathcal{L}_{\mathrm{logitsKD}} + \lambda_{2}\mathcal{L}_{\mathrm{featKD}} + \lambda_{3}\mathcal{L}_{\mathrm{assignKD}}.\nonumber
\end{equation}
% where $\mathcal{L}_{\mathrm{det}}$ is the original loss for DETR. 
where $\lambda_{1}=1, \lambda_{2}=20, \lambda_{3}=1$ are balancing weights of the proposed three KD loss terms. Since our distillation method follows the common DETR paradigm, it can be easily applied to a variety of detectors in DETR-families.
%-------------------------------------------------------------------------

\section{Experiments}

\subsection{Experiment Setup and Implementation Details}
\noindent\textbf{Dataset.} The challenging large-scale MS COCO benchmark~\cite{lin2014microsoft} is used in this study, where the $train2017$ (118K images) is utilized for training and $val2017$~(5K images) is used for validation. 
We use the standard COCO-style measurement, $i.e.$, average precision (mAP) for evaluation.

\noindent\textbf{DETR Models.}
% Furthermore, we conduct experiments on 
Three different DETR detectors are evaluated, including Deformable DETR~\cite{zhu2020deformable}, Conditional DETR~\cite{meng2021conditional}, and AdaMixer~\cite{gao2022adamixer}. We choose these three models for their representative model framework and their excellent performance. In the ablation study, we choose AdaMixer as the baseline for ablation and analysis for its easy training and quick convergence.
% because it converges quickly and is easy to train. 

\noindent\textbf{Implementation Details.}
Our codebase is built on MMdetection toolkit~\cite{chen2019mmdetection}. 
All models are trained on 8 NVIDIA V100 GPUs. Unless otherwise specified, we train the teacher model for 1$\times$ schedule (12 epochs) or 50 epochs using ResNet-101~\cite{he2016deep} as the backbone with Adam optimizer and train the student model with the same learning schedule using ResNet-50~\cite{he2016deep} as the backbone following each baseline's settings.

% DETRDistill has only a few hyper-parameters since the distillation strategy conforms to the nature of the detector itself and reuses many ready-made loss functions in DETR. Specifically, $\lambda_{1}=1, \lambda_{2}=20, \lambda_{3}=1$ are used for all detectors.

\begin{table}[t]
\begin{center}
\resizebox{\linewidth}{!}{
\begin{tabular}{c |l |c |cccc}
  \toprule
  Detector & Setting & Epoch & AP & AP$_S$ & AP$_M$ & AP$_L$ \\ 
  \midrule
  \midrule
  \multirow{6}*{\makecell[c]{AdaMixer \\ (100 Queries)}} & \cellcolor{gray!40}Teacher & 12 \cellcolor{gray!40}& 43.6 \cellcolor{gray!40}& 25.4\cellcolor{gray!40} & 46.8\cellcolor{gray!40} & 60.7\cellcolor{gray!40} \\   
   & Student\cellcolor{gray!40}& 12\cellcolor{gray!40} & 42.3 \cellcolor{gray!40} & 25.3\cellcolor{gray!40} & 44.8\cellcolor{gray!40} & 58.2\cellcolor{gray!40} \\
   & FGD~\cite{yang2022focal} & 12 & 40.7(-1.6)  & 23.4 & 43.3 & 55.8 \\
   & MGD~\cite{yang2022masked} & 12 & 42.3(+0.0)  & 24.5 & 45.0 & 58.9 \\ 
   & FitNet~\cite{romero2014fitnets} & 12 & 42.9(+0.6) & 24.7& 45.8& 59.4 \\
   & \textbf{Ours} & 12 & \textbf{44.7}(+2.4)  & \textbf{26.7} & \textbf{47.6} & \textbf{61.0} \\
  \midrule
  \multirow{6}*{\makecell[c]{Deformable DETR \\ (300 Queries)}} & Teacher\cellcolor{gray!40} & 50 \cellcolor{gray!40}& 45.5\cellcolor{gray!40} & 27.5 \cellcolor{gray!40}& 48.7\cellcolor{gray!40} & 60.3\cellcolor{gray!40}\\
   & Student\cellcolor{gray!40}& 50\cellcolor{gray!40} & 44.1 \cellcolor{gray!40} & 27.0 \cellcolor{gray!40}& 47.4 \cellcolor{gray!40}& 58.3\cellcolor{gray!40} \\
   & FGD~\cite{yang2022focal} & 50 & 44.1(+0.0)  & 25.9 & 47.7 & 58.8 \\
   & MGD~\cite{yang2022masked} & 50 & 44.0(-0.1) & 25.9 & 47.3 & 58.6 \\ 
   & FitNet~\cite{romero2014fitnets} & 50 & 44.9(+0.8) & 27.2& 48.4& 59.6 \\
   & \textbf{Ours} & 50 & \textbf{46.6}(+2.5)  & \textbf{28.5} & \textbf{50.0} & \textbf{60.4} \\
   \midrule
  \multirow{6}*{\makecell[c]{Conditional DETR \\ (300 Queries)}} & Teacher\cellcolor{gray!40} & 50\cellcolor{gray!40} & 42.4\cellcolor{gray!40} & 22.6 \cellcolor{gray!40}& 46.0\cellcolor{gray!40} & 61.2\cellcolor{gray!40}\\
   & Student\cellcolor{gray!40}& 50\cellcolor{gray!40} & 40.7\cellcolor{gray!40} & 20.3\cellcolor{gray!40} & 43.8\cellcolor{gray!40} & 60.0\cellcolor{gray!40} \\
   & FGD~\cite{yang2022focal} & 50 & 40.4(-0.3)  & 19.7 & 43.8 & 59.5 \\
   & MGD~\cite{yang2022masked} & 50 & 41.2(+0.5) & 20.6 & 44.6 & 60.5 \\ 
   & FitNet~\cite{romero2014fitnets} & 50 & 41.0(+0.3) & 20.2& 44.3& 59.9 \\
   & \textbf{Ours} & 50 & \textbf{42.9}(+2.2) & \textbf{21.6} & \textbf{46.5} & \textbf{62.2} \\
   \bottomrule 
\end{tabular}}
\end{center}
\caption{
% A comparison between our DETRDistill with existing state-of-the-art distillation methods.
Results for distillation on identical number of encoder and decoder stages for teacher and student network.
}
% }
% \vspace{1em} 
\label{tab:main_results}
\end{table}

\subsection{Main Results}
In this section, we mainly present experimental results on two kinds of settings: (1) identical number of encoder and decoder stages for teacher and student; (2) various numbers of encoder and decoder stages for teacher and student. In the \textit{supplementary material}, we further provide more experimental results on other settings to demonstrate the effectiveness of our approach, such as distillation to lightweight backbones and self-distillation.
% We compare our proposed DETRDistill with other state-of-the-art knowledge distillation methods on three representative DETRs: AdaMixer~\cite{gao2022adamixer}, Deformable DETR~\cite{zhu2020deformable} and Conditional DETR~\cite{meng2021conditional}.  
% For AdaMixer, the teacher and student models use 100 queries for 1$\times$ schedule training, and all other settings are kept consistent with the original paper~\cite{gao2022adamixer}.
% For Deformable DETR and Conditional DETR, the teacher and student models use 300 queries for 2$\times$ training and adopt the same setting except for the backbone.

\smallskip
\noindent\textbf{Distillation on identical number of encoder and decoder stages.}
The results are presented in Tab.~\ref{tab:main_results}. We can find that FitNet~\cite{romero2014fitnets} brings stable performance gains on all the DETR variants, while 
% the latest distillation methods for feature levels like 
MGD~\cite{yang2022masked} and FGD~\cite{yang2022focal} cannot work well for Transformer-based detectors and even may cause degradation of the results. 
% We infer the reason that AdaMixer adaptively mixes sampled features in the channel and spatial dimension, which conflicts with FGD. 
% Since the classical distillation methods are built upon CNN networks, we implement them on the feature maps from the neck behind the backbone instead of the encoder. 
However, it is obvious that our approach significantly enhances the student's performance and surpasses all other methods by a large margin on various detectors. Specifically, DETRDistill gains 2.4 AP on the AdaMixer, 2.5 AP on the Deformable DETR, and 2.2 AP on the Conditional DETR, which validates the effectiveness of our approach.

\begin{table}[h]
\begin{center}
\resizebox{\linewidth}{!}{
\begin{tabular}{c|c|cccc}
  \toprule
  \#Enc./Dec. & Setting & AP & AP$_S$ & AP$_M$ & AP$_L$  \\  
  \midrule
  \midrule
  \multirow{2}*{6/6}&\cellcolor{gray!40} Student &\cellcolor{gray!40} 33.0 &\cellcolor{gray!40} 13.8 &\cellcolor{gray!40} 35.1 &\cellcolor{gray!40} 50.9 \\
  & \textbf{Ours} & \textbf{38.1}(+5.1)  & \textbf{18.2} & \textbf{41.0} & \textbf{58.2} \\
  \midrule
  \multirow{2}*{6/3}&\cellcolor{gray!40} Student &\cellcolor{gray!40} 30.3 &\cellcolor{gray!40} 12.7 &\cellcolor{gray!40} 32.5 &\cellcolor{gray!40} 46.8 \\
  & \textbf{Ours} & \textbf{36.6}(+6.3)  & \textbf{15.2} & \textbf{39.8} & \textbf{56.0} \\
  \midrule
  \multirow{2}*{3/6}&\cellcolor{gray!40} Student &\cellcolor{gray!40} 33.0  &\cellcolor{gray!40} 13.9 & \cellcolor{gray!40}35.9 &\cellcolor{gray!40} 49.0 \\
  & \textbf{Ours} & \textbf{37.7}(+4.7)  & \textbf{17.4} & \textbf{40.8} & \textbf{57.3} \\
  \midrule
  \multirow{2}*{3/3}& \cellcolor{gray!40}Student &\cellcolor{gray!40} 29.9 & \cellcolor{gray!40}12.8 &\cellcolor{gray!40} 32.2 &\cellcolor{gray!40} 45.6 \\
  & \textbf{Ours} & \textbf{35.7}(+5.8)  & \textbf{14.4} & \textbf{38.7} & \textbf{53.8} \\
  
  \bottomrule
\end{tabular}}
% }
\end{center}
\caption{
Results for distillation on a various number of encoder and decoder stages for teacher and student network. The performance of the teacher model is 36.2~AP.
% Distillation on different Transformer Layers: Compressing the number of encoder layers and decoder layers from the same teacher with 36.2~AP.
}
\label{tab:enc_dec}
\end{table}

% \subsection{Distilling to Lightweight Backbones}

% Knowledge distillation aims to transfer as much knowledge as possible from a large model to a lightweight model for deployment on edge. With this in mind, we also apply DETRDistill with small backbones including ResNet-18 and MobileNetV2~\cite{sandler2018mobilenetv2} on the above DETRs. 
% % For AdaMixer, the teacher and student models use 100 queries for 1$\times$ training, and all other settings are kept consistent with the paper. For Deformable DETR and Conditional DETR, the teacher and student models adopt the same 2x training setting with 300 queries except for the backbone. 
% The results are shown in Tab.~\ref{tab:lightweight}. Our distillation method has achieved the best performance on both backbones. We achieve 2.0/2.6 mAP improvements on AdaMixer, 3.3/3.5 mAP on Deformable DETR, and 2.1/2.7 mAP on Conditional DETR, respectively.

% \input{tables/lightweight}   蒸轻量backbone

% \subsection{Self-Distillation}

% Self-distillation is a special case of knowledge distillation where the teacher and student models are aligned, with the only aim of improving the model's performance. The teacher and student all use ResNet-50 as the backbone. We compare the self-distillation performance of our method with FGD~\cite{yang2022focal} and MGD~\cite{yang2022masked} based on the above DETRs. Tab.\ref{tab:self_distill} shows that our DETRDistill gains 1.3 mAP, 2.3 mAP, and 1.8 mAP over the baselines. In contrast, FGD and MGD do not bring any improvement and even cause a decline in results.

\begin{figure}
  \centering
   \includegraphics[width=1.\linewidth]{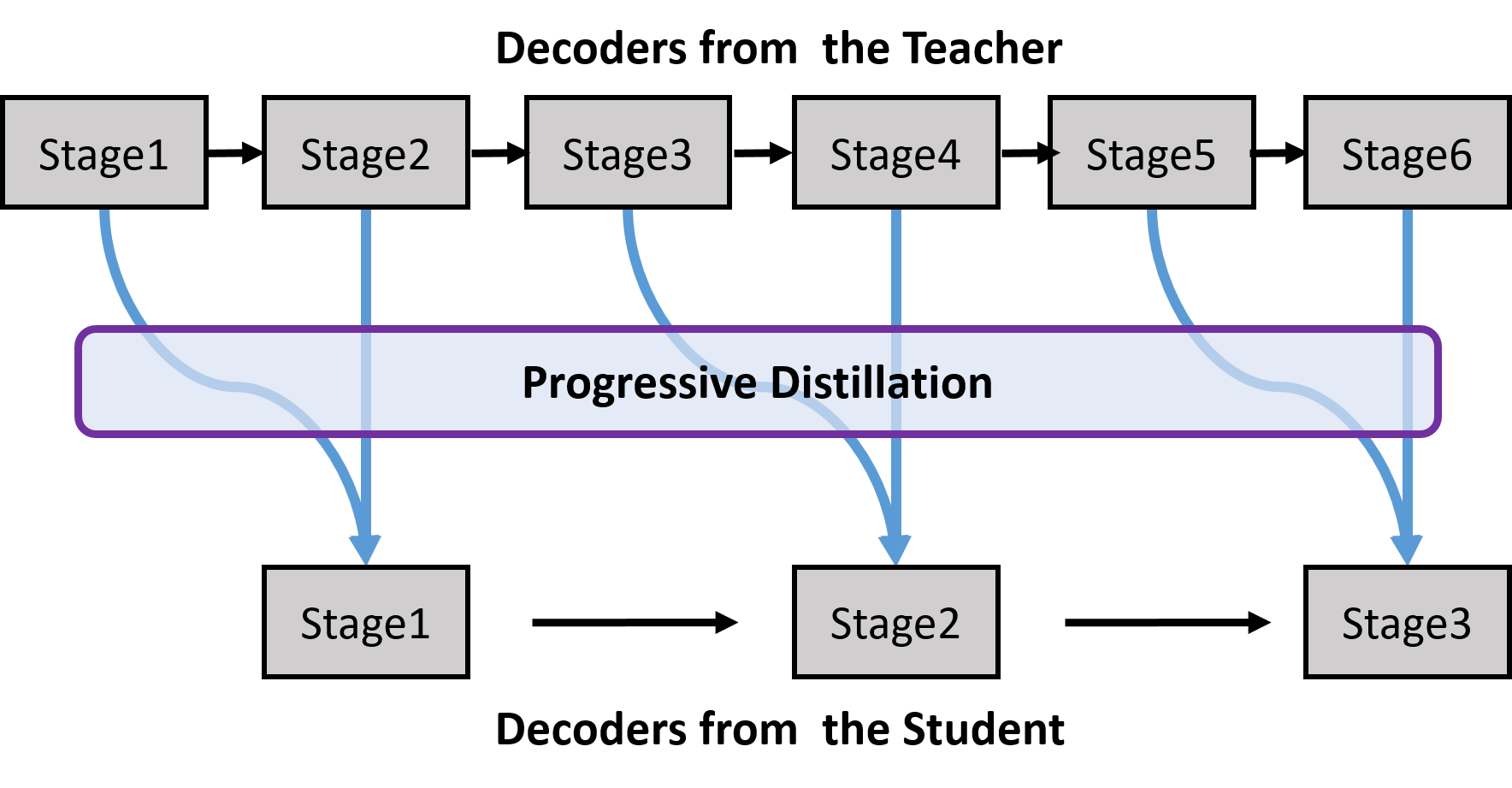}

   \caption{Distillation diagram with the different number of decoders.}
   \label{fig:tea6_stu3}
\end{figure}

\smallskip
\noindent\textbf{Distillation on a different number of encoders and decoders stages.}
Since the student model is usually smaller than the teacher model, there is no guarantee that both teacher and student have the same number of transformer encoder and decoder stages. Therefore, we also conduct experiments on the student model with a reduced number of encoders/decoders than that of teachers and explore the critical factors for them.
% Considering that some DETR variants have ample parameters in transformer layers, we also conduct experiments on the student model with a reduced number of encoders/decoders and explore the critical factors for them.
% In addition to the compression on backbones, some DETR variants have ample parameters in transformer layers, which cannot be ignored for the DETR distillation. So we construct some student models by reducing the number of encoders and decoders and exploring the critical factor for them. 
For this experiment, we choose Conditional DETR with ResNet-101 as the teacher model and ResNet-50 as the student model. All models are trained with the 1$\times$ schedule (12 epochs). The default number of encoders and decoders in the teacher model is 6 and we decrease the number of encoders and decoders from 6 to 3 for the student model.
% It is worth mentioning that when the number of decoders in the student is less than that in the teacher model, 
Since the default progressive distillation proposed in Sec.~\ref{sec:4.1} assumes that teacher and student have the same number of decoder stages, we simply group the teacher's decoders and make them follow our progressive strategy as shown in Fig.~\ref{fig:tea6_stu3} to handle the stage mismatch problem.\footnote{Please note that Target-aware Feature Distillation presented in Sec.~\ref{sec:feat_distill} is performed on the output of transformer encoders, which with the same resolutions as the input. Thus not affected by the change of encoder stages.}

% We decrease the original number of encoders and decoders from 6 to 3. 
The main results are shown in Tab.~\ref{tab:enc_dec}. 
It can be seen that decreasing the number of transformer encoders will not have a significant impact on the performance, while a reduction in the number of decoders leads to serious performance degradation.
However, our proposed DETRDistill significantly compensates for the performance gap. For example, with only 3 encoder and decoder stages, our approach helps the student model achieve 35.7 AP which is superior to the student model with 6 encoder and decoder stages without KD.
% gains 5.1 mAP and 4.7 mAP over the 6/6 and 3/6 Enc/Dec baselines, respectively. For the 3-decoder student model, our improvements yield much higher and reach 6.3 mAP and 5.8 mAP.

\section{Ablation Studies}
In this section, we are interested in ablating our approach from the following perspectives.

% \smallskip
\noindent\textbf{Effects of each component.}
To study the impact of each component in DETRDistill, we report the performance of each module in Tab.~\ref{tab:main_abl}. Our baseline starts from 42.3 AP. When logits-level distillation, feature-level distillation, and query-prior assignment distillation are applied separately, we can obtain the gain of 1.4 AP, 1.2 AP, and 0.6 AP, respectively. 
% The three components have different emphases when transferring teachers' knowledge. 这句话没有连贯性
Finally, the AP performance achieves 44.7 when all three modules are applied together, gaining a 2.4 AP absolute improvement.

\begin{table}[t]

\begin{center}

\resizebox{\linewidth}{!}{
\begin{tabular}{c|cccc}
  \toprule
  Distillations & AP  & AP$_S$ & AP$_M$ & AP$_L$  \\ 
  \midrule
  \midrule
   None & 42.3  & 25.3 & 44.8 & 58.2  \\
   % \hdashline
   LD & 43.7\textbf{(+1.4)}  & 25.3 & 46.5 & 60.7  \\  
   FD & 43.5\textbf{(+1.2)}  & 25.4  & 46.7 & 60.0   \\
   AD & 42.9\textbf{(+0.6)}  & 24.5  & 45.9 & 59.3 \\
   LD + FD & 44.3\textbf{(+2.0)}  & 25.8  & 47.0 & 61.0  \\
   LD + FD + AD  & 44.7\textbf{(+2.4)} & 26.7  & 47.6 & 61.0  \\
  \bottomrule
\end{tabular}}

\end{center}
\caption{Ablation study of each component in DETRDistill. LD stands for Hungarian-matching Logits Distillation; FD stands for Target-aware Feature Distillation; AD stands for Query-prior Assignment Distillation. 
% Results are reported on AdaMixer-R50 with standard 1 $\times$ schedule setting. \haiming{write performance gain}
}
\label{tab:main_abl}
\end{table}

\smallskip
\noindent\textbf{Selection of logits distillation.}
% \noindent\textbf{Ablations on samples for classification and regression branches}
Our default logits distillation is performed on positive classification predictions, positive and negative box regressions. We ablate various selections of these terms in Tab.~\ref{tab:pos_neg}. 
% In this section, we explore the performance of the logit distillation with positive/negative samples on the regression/classification branch, respectively. 
% We pre-define the maximum classification probability of the teacher's prediction as less than 0.1 as the background (negative). 
% The experimental results are shown in Table.~\ref{tab:pos_neg}. 
We can find that only distilling on positive predictions can only bring minor performance gains. While the distillation on negative box regression significantly improves the performance. This clearly verifies the importance of the proposed negative location distillation proposed in Sec.~\ref{sec:4.1}.
% using regression position and classification probability alone as extra supervision improves 1.3 mAP and 0.3 mAP, respectively. 
% This indicates that the location information is more challenging to learn for the DETRs, making it more critical in the distillation phase.
% Furthermore, we explore the effectiveness of positive and negative distillation. Table.~\ref{tab:pos_neg} shows that the negative location information can bring more gain (1.1 mAP \textit{vs} 0.3 mAP) than the positive location information. More detailed analysis of this is in~\cref{sec:4.1}.

\begin{table}[t]
\begin{center}
\resizebox{\linewidth}{!}{
\begin{tabular}{ccc|cccc}
\toprule
  Pos. Cls. & Pos. Reg. & Neg. Reg. & AP & AP$_S$ & AP$_M$ & AP$_L$  \\  
  \midrule
  \midrule
  % \multicolumn{3}{c|}{None} & 42.3  & 25.3 & 44.8 & 58.2 \\
  - & - & - & 42.3  & 25.3 & 44.8 & 58.2 \\
  % \hdashline
  % \hline
  \checkmark & & & 42.6 & 24.9 & 45.4& 58.9 \\ 
   &  \checkmark & & 42.6  & 24.4 & 45.6 & 58.8 \\
   &    & \checkmark& 43.4 & 25.2 & 46.2 & 60.1\\
   \checkmark & \checkmark & & 42.5 & 24.9 & 45.4 & 58.3\\
   &   \checkmark & \checkmark& 43.6 & 25.2 & 46.3 & 60.6\\
   \checkmark& \checkmark & \checkmark& \textbf{43.7} & \textbf{25.3} & \textbf{46.5} & \textbf{60.7} \\
  \bottomrule
\end{tabular}}
\end{center}
\caption{Ablations on the logits selection for distillation.
% loss of progressive distillation for classification and regression.
}
\label{tab:pos_neg}
\end{table}

\smallskip
\noindent\textbf{Necessary of progressive distillation.}
As presented in Sec.~\ref{sec:4.1}, we choose to progressively distill the knowledge  stage-by-stage from the teacher model to the student model, but not to only use the last stage output of the teacher model. Tab.~\ref{tab:progressive} shows that the former choice performs better.

% To verify the effectiveness of progressive distillation, we additionally explore two ways of transferring information. 
% The first is to regard the teacher's final predictions as the student's goal for all stages, represented by direct distillation. 
% As we can see from Table.~\ref{tab:progressive}, progressive instance distillation can achieve 0.6 mAP higher than direct distillation. 
% However, progressive distillation may cause the same query of a student to be assigned to different teacher queries at different stages.

% Based on this, we propose another learning approach: progressive guided distillation, which only performs bipartite graph matching in the last stage. Then, we use the matching orders as the correspondence between students' and teachers' queries in the previous stages. The experiment shows that this strategy causes a sharp decline (1.0 mAP), indicating that it is better to let students choose what they learn on demand at different stages rather than forcing.

\begin{table}[t]
\begin{center}
\resizebox{\linewidth}{!}{
\begin{tabular}{c|cccc}
  \toprule
  Strategies & AP  & AP$_S$ & AP$_M$ & AP$_L$  \\  
  \midrule
  \midrule
  Baseline & 42.3 & 25.3 & 44.8 & 58.2 \\
  Last-stage Distillation & 43.1 & \textbf{25.4} & 45.9 &59.7    \\ 
  Progressive Distillation & \textbf{43.7}  & 25.3 & \textbf{46.5} & \textbf{60.7}    \\
  \bottomrule
\end{tabular}}
% }
\end{center}
\caption{Ablation study on the necessity of progressive distillation. Last-stage distillation means only using the teacher's last decoder stage output to supervise all decoder stages of the student model.
% Comparison of different imitation learning approaches for instance distillation.
}
\label{tab:progressive}
\end{table}

\begin{figure}
  \centering
   \includegraphics[width=1.\linewidth]{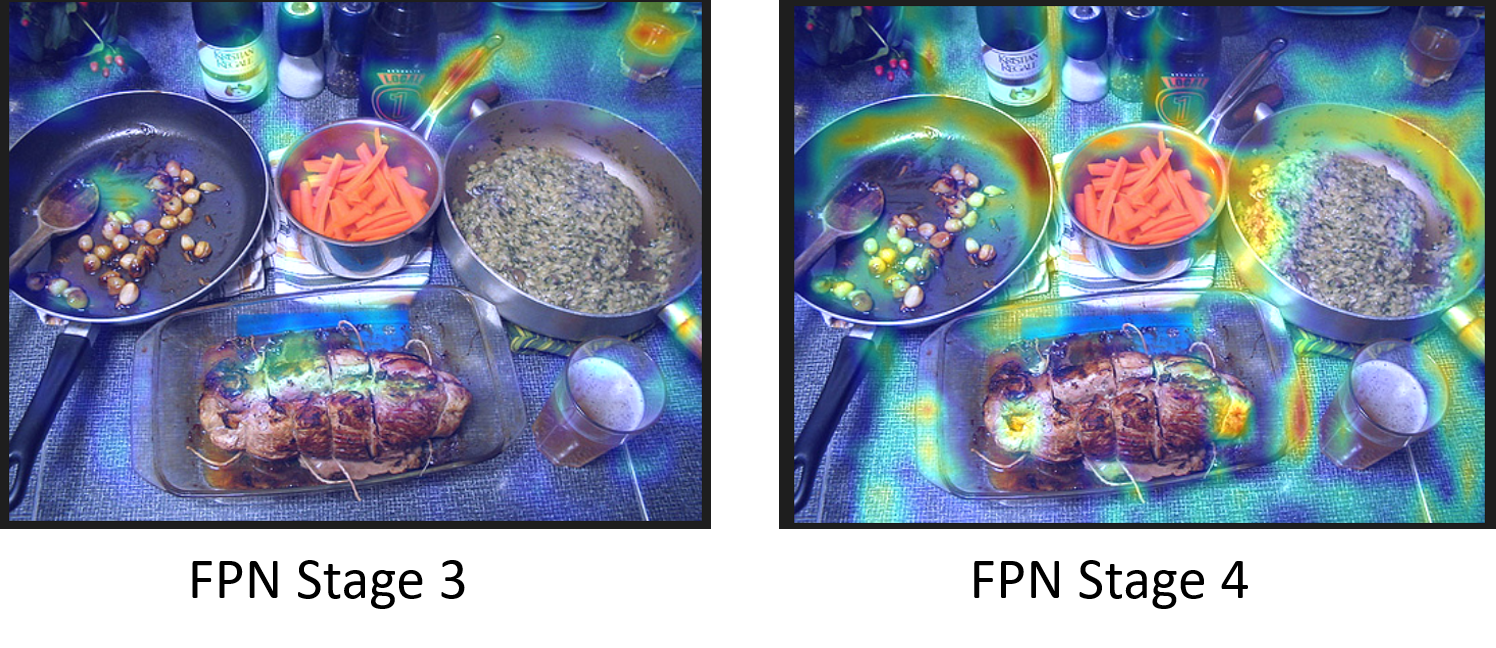}

   \caption{Visualization of selection mask on different FPN stages.}
   \label{fig:fpn_masks}
\end{figure}

\smallskip
\noindent\textbf{Visualization of selection mask in feature level distillation.}
% \noindent\textbf{Analysis on Distillation mask in feature-level}
In Sec.~\ref{sec:feat_distill}, we have visualized some query-based selection mask $\psi_i$ in Fig.~\ref{fig:query_mask}. Here, we are also interested in the quality weighted mask $\sum_i^M (q_{i}\cdot \psi_i)$ and Fig.~\ref{fig:fpn_masks} presents two visualizations of this mask on different FPN stages. 
% We visualize the soft mask in Fig.~\ref{fig:fpn_masks}, and 
It can be seen that the soft masks are activated near the object targets and retain certain contextual information, which is in line with our expectations. We also find that the stages with different strides in FPN have different knowledgeable regions from the mask.
% Moreover, it inspires us to discard the selection by ground truth boxes and should distill the feature on the fine-grained pixel-level.
Moreover, it inspires us to not use ground truth for selection mask generation but to use an immediate query-based selection mask for knowledge distillation as proposed in Sec.~\ref{sec:feat_distill}.

\smallskip
\noindent\textbf{Visualization of instability between predictions from different decoder stages.}
% \noindent\textbf{Analysis on Query-Prior Assignment Distillation}
In Sec.~\ref{sec:query_prior_assign_distill}, we have proposed a Query-prior Assignment Distillation to speed and stabilize the training of the student model. Here, we present an analysis of the instability between predictions from different decoder stages to investigate the effectiveness of the proposed distillation module. We utilize the same metric proposed by~\cite{li2022dn} to evaluate bipartite assignment's instability. However, our focus is on the stability of matching between different decoder stages, not between different epochs. The predictions in the $k$-th decoder stage are denoted as $\mathbf{O}^{\mathbf{k}}=\left\{O_{0}^{k}, O_{1}^{k}, \ldots, O_{N-1}^{k}\right\}$, where $N$ is the number of predicted objects, and the GTs are denoted as $\mathbf{T}=\left\{T_{0}, T_{1}, \ldots, T_{M-1}\right\}$ where $M$ is the number of ground truth targets. Then we compute the index vector $\mathbf{V}^{\mathbf{k}}=\left\{V_{0}^{k}, V_{1}^{k}, \ldots, V_{N-1}^{k}\right\}$ to store the assignment result of $k$-th decoder stage as
\begin{equation}\label{8}
V_{n}^{k}=\left\{\begin{array}{ll}
m, & \text { if } O_{n}^{k} \text { matches } T_{m} \\
-1, & \text { if } O_{n}^{k} \text { matches nothing }\nonumber
\end{array}\right.
\end{equation}
And the instability ($IS$) between decoder stage $k$ and stage $k+1$ can be calculated as:
\begin{equation}\label{8}
IS^k=\sum_{j=0}^{N}\mathds{1}({V_{n}^{k} \neq V_{n}^{k+1}}) \nonumber
\end{equation}

\begin{figure}
  \centering
   \includegraphics[width=1.\linewidth]{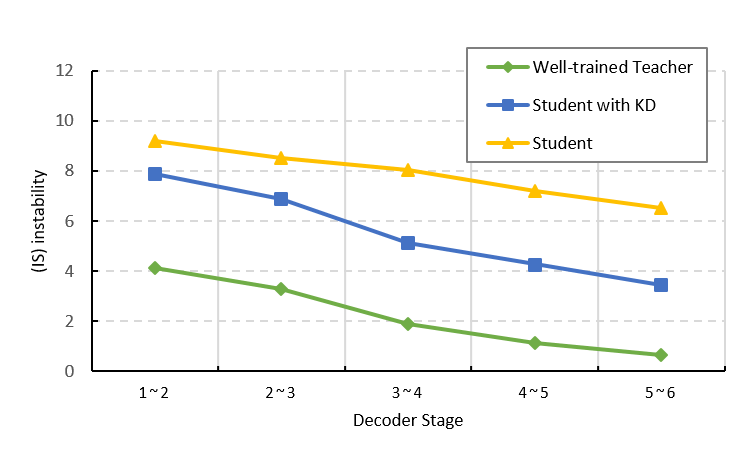}

   \caption{The instability ($IS$) of teacher, student, and student with our Query-prior Assignment Distillation after training one epoch.
   % between different decoder stages during the prophase training.
   }
   \label{fig:stability}
\end{figure}

We calculate the $IS$ metric on the well-trained teacher model and the student baseline model. As for the model with our Query-prior Assignment Distillation, we compute the $IS$ metric on the immediate checkpoint of the first training epoch. As Fig.~\ref{fig:stability} shows, the naive student has higher instability than the teacher model. But the $IS$ score decreases a lot when using the proposed distillation module for training only one epoch, which clearly verifies that such a distillation module can help the student model train more stably and thus speed up the convergence rate.
% and the random initialization model in the prophase phase, as shown in the Fig.~\ref{fig:stability}. It can be seen that applying our Query-Prior Assignment Distillation can effectively improve the stability of the student's query matching.

\section{Conclusion}
This paper introduces a universal knowledge distillation framework for DETR-style detectors named DETRDistill. Our method contains three distillation modules: Hungarian-matching Logits Distillation, Target-aware Feature Distillation, and Query-prior Assignment Distillation. Extensive experiments on the competitive COCO benchmark demonstrate the effectiveness and generalization of our approach. Notably, it achieves considerable gains on various transformer-based detectors compared with current state-of-the-art knowledge distillation methods. We hope that DETRDistill can serve as a solid baseline for DETR-based knowledge distillation for future research.

{\small
\bibliographystyle{ieee_fullname}
\bibliography{egbib}
}

\end{document}